\documentclass[sigconf,screen]{acmart}
\renewcommand\footnotetextcopyrightpermission[1]{}
\usepackage{algorithm}
\usepackage{algorithmic}
\usepackage{multirow}
\usepackage{array}
\usepackage{graphicx}
\usepackage{booktabs}
\usepackage{makecell}

\AtBeginDocument{%
  }

\setcopyright{acmlicensed}
\copyrightyear{2018}
\acmYear{2018}
\acmDOI{XXXXXXX.XXXXXXX}
\acmConference[]{}{}{}
\acmISBN{978-1-4503-XXXX-X/2018/06}

\acmSubmissionID{3409}

\begin{document}
\setlength{\abovedisplayskip}{6pt}
\setlength{\belowdisplayskip}{6pt}
\setlength{\abovedisplayshortskip}{6pt}
\setlength{\belowdisplayshortskip}{6pt}
\title{Breaking Watermarks in the Frequency Domain: A Modulated Diffusion Attack Framework}

\author{Chunpeng Wang}
\affiliation{%
  \institution{Qilu University of Technology
(Shandong Academy of Sciences)}
  \city{Jinan}
  \state{Shandong Province}
  \country{China}
}
\email{wangcp@qlu.edu.cn}

\author{Binyan Qu}
\affiliation{%
  \institution{Qilu University of Technology
(Shandong Academy of Sciences)}
  \city{Jinan}
  \state{Shandong Province}
  \country{China}
}
\email{10431240208@stu.qlu.edu.cn}

\author{Xiaoyu Wang}
\affiliation{%
  \institution{Dalian Maritime University}
  \city{Dalian}
  \state{Liaoning Province}
  \country{China}
}
\email{qluwxy@163.com}

\author{Zhiqiu Xia}
\affiliation{%
  \institution{Qilu University of Technology
(Shandong Academy of Sciences)}
  \city{Jinan}
  \state{Shandong Province}
  \country{China}
}
\email{xzqjsdtc@163.com}

\author{Shanshan Zhang}
\affiliation{%
  \institution{Nanjing University of Science and Technology}
  \city{Nanjing}
  \state{Jiangsu Province}
  \country{China}
}
\email{shanshan.zhang@njust.edu.cn}

\author{Yunan Liu}
\affiliation{%
  \institution{Shandong Jianzhu University}
  \city{Jinan}
  \state{Shandong Province}
  \country{China}
}
\email{liuyunan@njust.edu.cn}

\author{Qi Li}
\affiliation{%
  \institution{Qilu University of Technology
(Shandong Academy of Sciences)}
  \city{Jinan}
  \state{Shandong Province}
  \country{China}
}
\email{qluliqi@163.com}

\renewcommand{\shortauthors}{}

\begin{abstract}
  Digital image watermarking has advanced rapidly for copyright protection of generative AI, yet the comparatively limited progress in watermark attack techniques has broken the attack-defense balance and hindered further advances in the field. In this paper, we propose FMDiffWA, a frequency-domain modulated diffusion framework for watermark attacks. Specifically, we introduce a frequency-domain watermark modulation (FWM) module and incorporate it into the sampling stages both the forward and reverse diffusion processes. This mechanism enables selective modulation of watermark-related frequency components, thereby allowing FMDiffWA to effectively neutralize the invisible watermark signals while preserving the perceptual quality of the attacked watermarked images. To achieve a better trade-off between attack efficacy and visual fidelity, we reformulate the training strategy of conventional diffusion models by augmenting the canonical noise estimation objective with an auxiliary refinement constraint. Comprehensive experiments demonstrate that FMDiffWA achieves superior visual fidelity compared to existing watermark attacks, while exhibiting strong generalization across diverse watermarking schemes.
\end{abstract}




\maketitle

\section{Introduction}
With the rapid advances in generative AI and social media platforms, the dissemination of images and videos has become increasingly convenient. However, the widespread circulation of AI-synthesized visual content poses a serious threat to the authenticity and credibility of online information. Digital watermarking has been widely studied due to its strong capability for copyright protection\cite{chen2025learning,Wan2022ACS}. Research in this field has mainly focused on two directions: watermarking algorithms and watermark attacks. Watermarking algorithms are designed to improve the robustness and integrity of watermarks through optimized embedding strategies, whereas attack methods aim to eliminate the watermark signals via various attacks, thereby preventing their reliable extraction.

In the field of digital watermarking, watermarking algorithms have developed rapidly in recent years, evolving from traditional spatial-domain methods \cite{su2024efficient,faheem2023edge}, frequency-domain methods \cite{cao2024universal,urvoy2014perceptual,Shen2021ADB}, and orthogonal moment-based methods \cite{Tang2024ARR,He2024ExploringAI,Zhang2024FastIR,Fu2023RobustRW}, to deep learning-based methods \cite{Zhu2018HiDDeNHD,Jia2021MBRSER,DBLP:conf/aisec-ws/HuangYHSCC25,Luo2023IRWArtLW,Kou2025IWRNAR}. In contrast, researches on watermark attacks remain relatively underexplored. This imbalance in development offers a new perspective: stronger and more effective watermark attacks can enable a more comprehensive evaluation of watermark robustness, thereby pushing the continued improvement and evolution of watermarking algorithms.
\begin{figure}[htbp]
  \centering
  \includegraphics[page=1,width=1\linewidth]{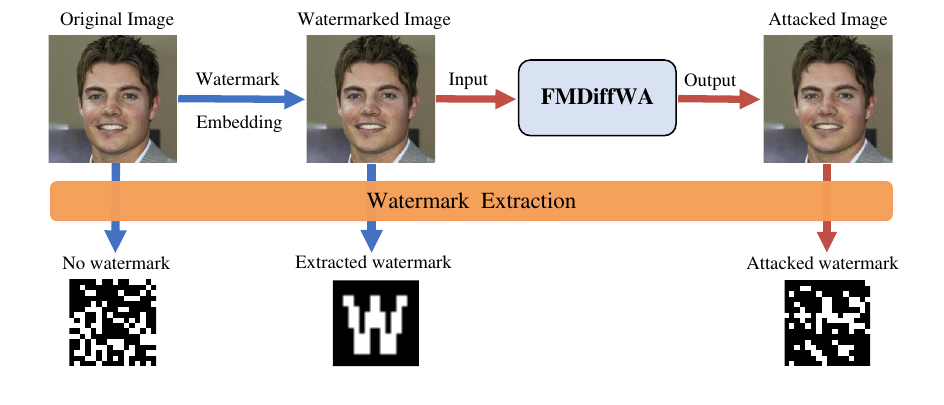}
  \caption{Overview of the proposed FMDiffWA}
  \label{Figure:Figure1}
\end{figure}
Conventional watermark attacks mainly fall into two categories: geometric attacks and image processing attacks. Geometric attacks modify the geometric structure of an image through operations such as scaling, rotation, and cropping, thereby disrupting the spatial synchronization of the watermark and making its detection and extraction more challenging. In contrast, image processing attacks attempt to eliminate watermark signals using common image editing operations, including compression, filtering, and noise injection. Recent advances in deep learning have stimulated the initial development of watermark attacks, as exemplified by CNN-based approaches \cite{Nam2020WANWA,Geng2020RealtimeAO,Hatoum2021UsingDL,Wang2024HIWANetAH} and GAN-based methods \cite{Li2022ConcealedAF}. However, this line of research remains in its infancy, since existing methods still exhibit limited attack capability, poor cross-scheme generalization, and difficulty in achieving a favorable trade-off between attack effectiveness and image quality.

To address the above issues, we propose FMDiffWA, a frequency-domain modulated diffusion framework for watermark attack. FMDiffWA leverages the conditional diffusion paradigm by taking the watermarked image as the input, while incorporating a random sampling mechanism to suppress irrelevant content artifacts. To accommodate images with varying resolutions, a patch-based processing strategy \cite{zdenizci2022RestoringVI} is integrated. Furthermore, we design a frequency-domain watermark modulation (FWM) module to better harness the original texture details preserved during forward diffusion, enabling precise Fourier-space manipulation of watermark features during the initial sampling stage. To mitigate the visual quality degradation inherent in standard noise estimation, we develop a two-stage training strategy. In the first stage, the model is optimized using the canonical noise estimation objective. In the second stage, we introduce an auxiliary refinement constraint by supervising the sampled results with corresponding watermark-free images, thereby significantly enhancing the clarity and fidelity of the generated outputs. The main contributions of this paper can be summarized as follows:
\begin{itemize}
\item We propose FMDiffWA, a frequency-domain modulated diffusion framework for watermark attack. FMDiffWA is built upon conditional diffusion, with random sampling for artifact suppression and patch-based processing for watermark attacks on images of varying resolutions.
\item We introduce a frequency-domain watermark modulation (FWM) module for precise watermark manipulation. FWM module enables fine-grained modulation of watermark frequency components by exploiting texture details preserved during forward diffusion.
\item We develop a two-stage training strategy that introduces an auxiliary refinement constraint supervised by watermark-free images, significantly improving the visual fidelity of attacked results while achieving a better trade-off between attack ability and image quality. 
\end{itemize}
\section{Related Work}
In this section, we review some related works to our proposed methods for watermark attack using diffusion model.
\subsection{Watermarking algorithm}
Watermarking algorithm aims to embed copyright information into host images while simultaneously satisfying two fundamental requirements: imperceptibility and robustness. Traditional watermarking algorithms can be broadly categorized into spatial-domain, frequency-domain, and orthogonal-moment-based approaches. Spatial-domain methods \cite{su2024efficient,faheem2023edge} directly modify pixel values to embed watermark information. Frequency-domain methods \cite{cao2024universal,urvoy2014perceptual,Shen2021ADB} hide watermark signals into image frequency coefficients to improve robustness against common attacks. Orthogonal-moment-based methods \cite{Tang2024ARR,He2024ExploringAI,Zhang2024FastIR,Fu2023RobustRW} exploit the inherent invariance of orthogonal moments to enhance the stability of watermarks under geometric and signal-processing transformations. In recent years, deep learning-based methods have further advanced watermarking performance by learning more expressive feature representations. Existing approaches mainly include end-to-end frameworks \cite{Zhu2018HiDDeNHD,Jia2021MBRSER,DBLP:conf/aisec-ws/HuangYHSCC25} and reversible network-based methods\cite{Luo2023IRWArtLW}. Zhu et al. \cite{Zhu2018HiDDeNHD} proposed Hidden, the first end-to-end watermarking framework, which introduces a noise layer to simulate various transformations during watermark embedding and extraction. Reversible network-based methods \cite{Luo2023IRWArtLW} preserve information via invertible transformations, thereby enabling more accurate watermark extraction. Kou et al. \cite{Kou2025IWRNAR} combined an end-to-end framework with reversible networks and further introduced a contrastive learning to improve the robustness and imperceptibility.
\subsection{Watermark attacks}
Watermark attacks are designed to disrupt, remove, or forge embedded watermarks, thereby hindering their accurate extraction. Traditional attacks typically degrade watermark signals through image processing operations like compression, filtering, and noise addition, but they inevitably impair the visual quality of the image. Deep learning has fundamentally advanced watermark attacks. While initial efforts focused on CNNs for rapid or denoising attacks \cite{Nam2020WANWA,Geng2020RealtimeAO,Hatoum2021UsingDL} the attention has shifted toward generative models. Specifically, GANs have been optimized with perceptual constraints to realize targeted attacks. More recently, diffusion models have established new state-of-the-art benchmarks, leveraging distance conditioning \cite{Li2023DiffWADM} or asymmetric loss functions \cite{Wang2024HIWANetAH} to faithfully reconstruct the watermarked-free image domain. Despite their success, diffusion-based pipelines remain mainly tied to spatial-level features, leaving the complementary reconstruction capability of frequency-domain features largely untapped. To address this limitation, we explicitly inject Fourier priors into the iterative sampling stage. This design enables more precise suppression of watermark-related features while maintaining the high-frequency structural details of the original image.
\subsection{Diffusion model}
Diffusion models\cite{Ho2020DenoisingDP,SohlDickstein2015DeepUL} have shown remarkable potential in image restoration tasks, including image inpainting \cite{DBLP:conf/nips/ShizXDPXH0F24}, deblurring \cite{Chen2025CDRMCD}, and super-resolution \cite{Jeong2025LatentSS,DBLP:conf/cvpr/ZhangYSG25}, owing to their powerful distribution modeling capability and flexible conditional guidance mechanism. By progressively refining corrupted observations toward the target distribution, diffusion models are particularly suited for tasks that require both precise local modification and global visual consistency. Recent studies have further demonstrated their effectiveness in restoration. Shi et al. \cite{DBLP:conf/nips/ShizXDPXH0F24} incorporated residual modules into a conditional diffusion model to improve the sampling efficiency. Chen et al. \cite{Chen2025CDRMCD} proposed a controllable diffusion restoration model that enables adaptive image deblurring through introducing controllable adjustment factors. Özdenizci et al. \cite{zdenizci2022RestoringVI} designed a block-based diffusion model for image restoration, allowing image reconstruction at arbitrary resolutions. The effectiveness of diffusion models in image restoration offers a strong basis for local editing and high-fidelity image reconstruction. Motivated by these advantages, a diffusion-based framework is adopted in this paper to achieve precise and visually imperceptible watermark attacks.

\section{Preliminaries}
\subsection{Watermark Embedding and Extraction}
A typical robust watermarking algorithm is intended to enable covert information embedding and reliable recovery, and its fundamental architecture generally comprises an encoder $E$ and a decoder $D$. In the embedding stage, the encoder takes the original image $I$ and watermark sequence $w$ as inputs and generates a watermarked image $I_w$, which maintains high visual consistency with $I$. In the extraction stage, the decoder is required to recover the watermark sequence $w$ from the possibly attacked watermarked image $I_w$. Formally, the above process can be expressed as:
\begin{equation}
I_w = E(I, w)
\end{equation}
\begin{equation}
w' = D\left( \Delta \left( I_w \right) \right)
\end{equation}
where, $\Delta$ is represents the allowable distortions or attacks that may affect the watermarked image. The watermark algorithm is considered robust if the decoder can correctly recover the same bit sequence $w'$ from the possibly distorted watermarked image $I_w$. 
\begin{figure*}[!t]
    \centering
    \includegraphics[page=2,width=1\textwidth]{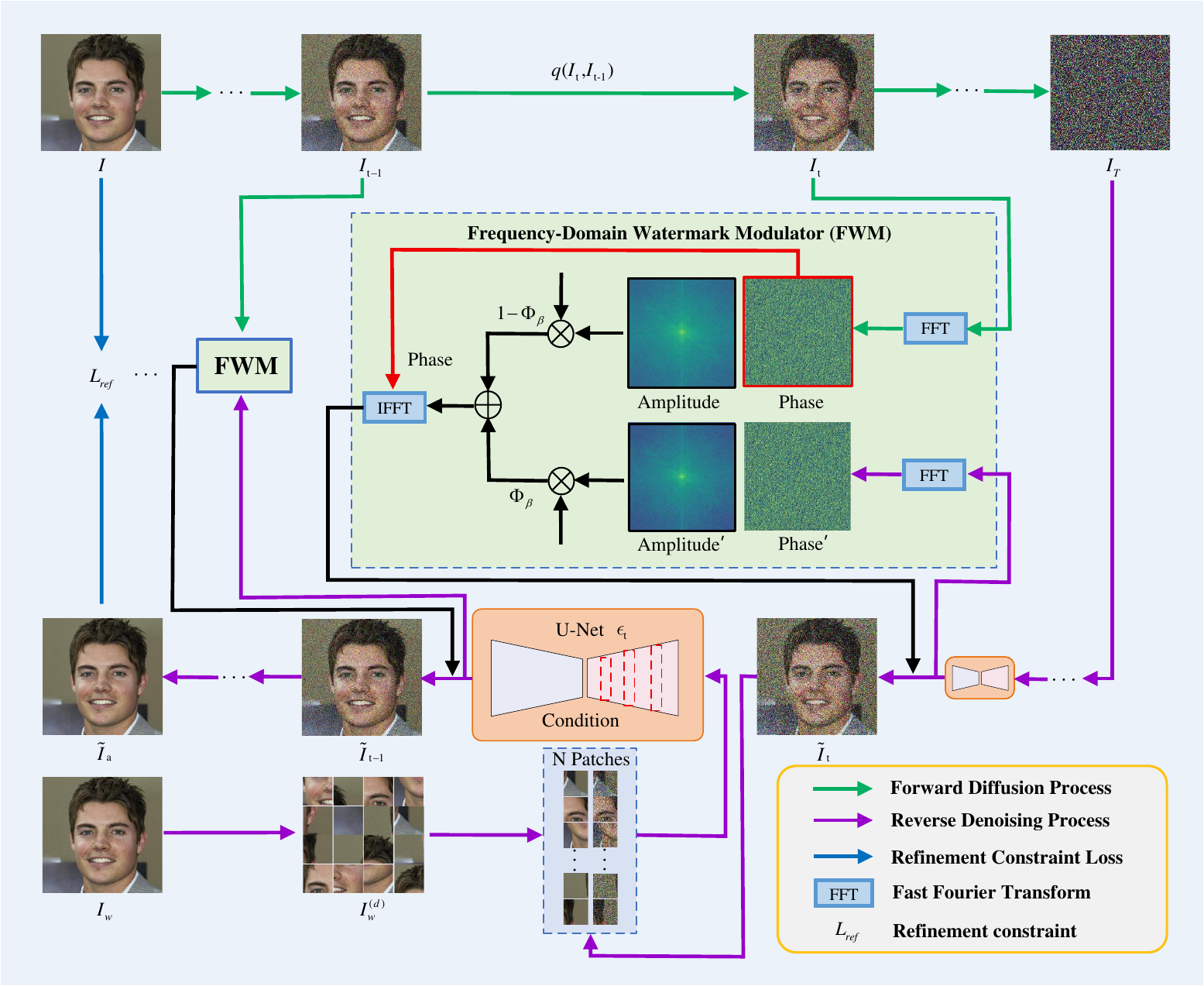}
    \caption{Frequency-domain modulated watermark attack module, including both training and sampling processes. The training stage consists of noise estimation and refinement constraints, while the sampling stage progressively transforms noise into a watermark-destroyed image guided under frequency-domain modulation.}
    \label{fig:Figure2}
\end{figure*}
\subsection{Diffusion Model Background}
Denoising Diffusion Probabilistic Models (DDPMs) \cite{Ho2020DenoisingDP} are latent-variable generative models that learn to generate data by reversing a progressive noising process. In general, a DDPM consists of two fundamental processes: a forward diffusion process and a reverse denoising process. The forward diffusion process progressively transforms an original image $I$ into Gaussian noise over $T$ steps. At each diffusion step, Gaussian noise is added to the output of the previous step $I_{t-1}$ as follows:
\begin{equation}
q(I_t | I_{t-1}) = \mathcal{N}\left(I_t; \sqrt{1 - \beta_t} I_{t-1}, \beta_t \omega\right)
\label{eq:forward_process}
\end{equation}
where $\beta_t$ denotes the variance of Gaussian noise and $I_t$ represents the noisy image at step $t$. $\omega$ is the identity matrix. The overall diffusion process forms a Markov chain, expressed as:
\begin{equation}
q(I_{1:T} | I_0) = \prod_{t=1}^T q(I_t | I_{t-1})
\label{eq:noise}
\end{equation}
where, $q(I_t | I_{t-1})$ can equivalently be written as $I_t = \sqrt{\alpha_t} I_{t-1} + \sqrt{1 - \alpha_t} \epsilon_{t-1}$, and $\alpha_t = 1 - \beta_t$. According to Markov property, $I_t$ at each step depends only on $I_{t-1}$. As the noise added at each step is independent, the diffusion process can be factorized into a product of conditional distributions, describing the transition from $I$ to $I_{1:T}$. By recursive expansion, it follows that:
\begin{equation}
I_t = \sqrt{\bar{\alpha}_t} I + \sqrt{1 - \bar{\alpha}_t} \epsilon_t
\end{equation}
where $\bar{\alpha}_t = \prod_{i=1}^t \alpha_i$, $\quad \epsilon_t \sim \mathcal{N}(0, I)$. The reverse denoising is formulated as a Markov chain initialized from $I_T$. A learned denoising function is then applied iteratively to progressively remove noise and reconstruct the target data:
\begin{equation}
p_\theta(I_{t-1} | I_t, I_w) = \mathcal{N}\left(I_{t-1}; \mu_\theta(I_t, I_w, t), \sigma_t^2 \omega\right)
\label{eq:eq6}
\end{equation}
where $\sigma_t^2 = \frac{1 - \bar{\alpha}_{t-1}}{1 - \bar{\alpha}_t} \beta_t$, $\quad 
\mu_\theta(I_t, I_w, t) = \frac{1}{\sqrt{\alpha_t}} \left( I_t - \frac{1 - \alpha_t}{\sqrt{1 - \bar{\alpha}_t}} \epsilon_\theta(I_t, I_w, t) \right)$, and $\epsilon_0(I_t, I_w, t)$ denotes the noise predicted by the denoising network. The loss function is formulated as: 
\begin{equation}
\mathcal{L}_{\text{noise}} = \mathbb{E}_{I, I_w, t, \epsilon_t} \left[ \left\| \epsilon_t - \epsilon_\theta\left( \sqrt{\bar{\alpha}_t} I + \sqrt{1 - \bar{\alpha}_t} \epsilon_t, I_w, t \right) \right\|^2 \right]
\label{eq:eq7}
\end{equation}

\section{Methodology}
The main idea of our proposed FMDiffWA is to investigate image watermark attacks from the perspective of diffusion models. FMDiffWA adopts a conditional diffusion model that takes the watermarked image as input and generates outputs aligned with the distribution of the watermark-free images. In this way, the watermark information can be effectively removed while maintaining high visual fidelity. To address the instability of watermark removal caused by the reliance on noise prediction in conventional diffusion models, a frequency-domain watermark modulation (FWM) module is introduced to explicitly exploit spectral information during both forward and reverse sampling. By modulating the phase and magnitude components in the Fourier domain, FWM effectively suppresses the embedded watermark features and significantly improves the attack imperceptibility. Moreover, we explore a semantic structure preservation mechanism in the forward diffusion stage, enabling more faithful recovery of the original image texture during reverse generation. In this section, we explicitly introduce the overall framework of FMDiffWA, followed by a detailed description of the proposed frequency-domain modulation module. Finally, a detailed description of the stage-wise training strategy is provided.
\begin{algorithm}[!t]
\caption{The Training Process }
\label{alg:alg1}
\begin{algorithmic}[1]
\small
\REQUIRE \hspace{-0.1cm}
Original image $I$, watermarked image $I_w$, denoiser network $\epsilon_\theta(I_t,I_w,t)$,
transitional iteration $K$, time steps $T$, sampling steps $S$, maximum perturbation $L_{\max}$, minimum perturbation $L_{\min}$, and the mask $\Phi_\beta$ in FWM.

\WHILE{not converged}
    \STATE Generate a random size binary mask $M_i$
    \STATE $I^{(i)} = \text{Crop}(M_i \circ I)$ and $I_w^{(i)} = \text{Crop}(M_i \circ I_w)$
    \STATE Sample random timestep $t \sim \text{Uniform}\{1, \dots, T\}$
    \STATE $\epsilon_t \sim \mathcal{N}(0, \mathbf{I})$

    \IF{Current iterations $\le K$}
        \STATE Take gradient descent step on 
        \STATE$  \nabla_\theta \left\| \epsilon_t - \epsilon_\theta\left( \sqrt{\bar{\alpha}_t}I^{(i)} + \sqrt{1-\bar{\alpha}_t}\epsilon_t, I_w^{(i)}, t \right) \right\|^2$
    \ELSE
        \STATE $\tilde{I}_{t_S}^{(i)} \sim \mathcal{N}(0, \mathbf{I})$
        \FOR{$j = S:1$}
            \STATE $t = (j-1)\cdot T/S + 1$
            \STATE $t_{\text{next}} = (j-2)\cdot T/S + 1$ if $j>1$ else $0$
            \STATE $\tilde{\epsilon}_t = \epsilon_\theta\left( \tilde{I}_t^{(i)}, I_w^{(i)}, t \right)$
            \STATE $I_{t_{\text{next}}}^{(i)} = \sqrt{\bar{\alpha}_{t_{\text{next}}}} I^{(i)} + \sqrt{1-\bar{\alpha}_{t_{\text{next}}}} \tilde{\epsilon}_{t_{\text{next}}}$
            \STATE $\tilde{I}_{t_{\text{next}}}^{(i)} = \sqrt{\bar{\alpha}_{t_{\text{next}}}} \left( \frac{\tilde{I}_t^{(i)} - \sqrt{1-\bar{\alpha}_t} \cdot \tilde{\epsilon}_t}{\sqrt{\bar{\alpha}_t}} \right) + \sqrt{1-\bar{\alpha}_{t_{\text{next}}}} \cdot \tilde{\epsilon}_t$
            \STATE $\tilde{I}_{t_{\text{next}}}^{(i)} = \text{FWM}\left( I_{t_{\text{next}}}^{(i)}, \tilde{I}_{t_{\text{next}}}^{(i)}, \Phi_\beta \right)$
            \STATE $L(t) = L_{\min} + (L_{\max} - L_{\min}) \cdot t / T$
            \STATE $\tilde{I}_{t_{\text{next}}}^{(i)} = \tilde{I}_{t_{\text{next}}}^{(i)} + L(t)$
        \ENDFOR
        \STATE Take gradient descent step on $\nabla_\theta \mathcal{L}_{\text{ref}}\left( \tilde{I}_a^{(i)}, I^{(i)} \right)$
    \ENDIF
\ENDWHILE
\RETURN $\theta$
\end{algorithmic}
\end{algorithm}

\begin{algorithm}[!t]
\caption{The Sampling Process }
\label{alg:alg2}
\begin{algorithmic}[1]
\small
\REQUIRE \hspace{-0.1em}
Watermarked image $I_w$, denoiser network $\epsilon_\theta(\tilde{I}_t,I_w,t)$, number of sampling steps $S$,total time steps $T$, and dictionary of $N$ overlapping patch locations.

\STATE $\tilde{I}_{t_S} \sim \mathcal{N}(0, \mathbf{I})$
\FOR{$j = S:1$}
    \STATE $t = (j-1)\cdot T/(S-1) + 1$
    \STATE $t_{\text{next}} = (j-2)\cdot T/(S-1) + 1$ \text{ if } $j>1$ \text{ else } $0$
    \STATE $\widehat{\Omega}_t = 0$ and $K = 0$
    \FOR{$d = 1:N$}
        \STATE $\tilde{I}_t^{(d)} = \text{Crop}\left(M_d \circ \tilde{I}_t\right)$ and $I_w^{(d)} = \text{Crop}\left(M_d \circ I_w\right)$
        \STATE $\widehat{\Omega}_t = \widehat{\Omega}_t + M_d \cdot \epsilon_\theta\left(\tilde{I}_t^{(d)}, I_w^{(d)}, t\right)$
        \STATE $K = K + M_d$
    \ENDFOR
    \STATE $\widehat{\Omega}_t = \widehat{\Omega}_t \oslash K$ \quad \{$\oslash$: element-wise division\}
    \STATE $\tilde{I}_{t_{\text{next}}} = \sqrt{\bar{\alpha}_{t_{\text{next}}}} \left( \frac{\tilde{I}_t - \sqrt{1-\bar{\alpha}_t} \cdot \widehat{\Omega}_t}{\sqrt{\bar{\alpha}_t}} \right) + \sqrt{1-\bar{\alpha}_{t_{\text{next}}}} \cdot \widehat{\Omega}_t$
\ENDFOR
\RETURN $\tilde{I}_a$
\end{algorithmic}
\end{algorithm}
\subsection{Overview}
The overall pipeline of watermark embedding, attack, and extraction are presented in Fig. \ref{Figure:Figure1}. Specifically, watermarks are first embedded into images using various watermarking algorithms. Subsequently, the proposed FMDiffWA is applied to attack the watermarked images. Finally, the attacked watermarked images fed into the corresponding extraction algorithms to extract the watermarks. Fig. \ref{fig:Figure2} details the implementation of the watermark attack module, with its key procedures summarized in Algorithm \ref{alg:alg1} and Algorithm \ref{alg:alg2}. As shown in Fig. \ref{fig:Figure2}, a grid-based parsing scheme \cite{zdenizci2022RestoringVI} is adopted to randomly partition the input image into overlapping patches. These patches are then merged with corresponding positional dictionaries, and the final noise prediction is obtained by averaging the estimates over overlapping pixels. 
\subsection{Frequency-domain Watermark Modulation Module}
When DDPMs are applied to image reconstruction, the reverse process relies primarily on the predicted noise, whereas the forward diffusion process may retain rich fine-grained information that is valuable for watermark attack. To more effectively integrate the forward and reverse processes for watermark attack, a frequency-domain watermark modulation (FWM) module is designed. FWM module applies the Fourier transform to decompose an image into its magnitude and phase components. the magnitude component captures low-level statistical features of the input image and may contain substantial watermark-related information, whereas the phase component primarily encodes high-level semantic structures, including image geometry and texture details. Based on this observation, both magnitude and phase information preserved in the forward diffusion process are exploited to provide informative guidance for the sampling stage. By mining structural, color, and watermark-free image cues from the forward process, the proposed FWM module facilitates more effective disruption of watermark embeddings and promotes the generation of realistic, high-fidelity images.

The details of FWM module are illustrated the green box of Fig. \ref{fig:Figure2}. Noise is added to the image according to Eq. \eqref{eq:forward_process}, and image reconstruction is performed via sampling based on Eq. \eqref{eq:eq6}. During sampling, the Fourier transform is applied to decompose images $I_{t-1}$ and $\tilde{I}_{t-1}$ into the corresponding magnitude and phase components \cite{Jiang2024WhenFF}. A Fourier transform is adopted independently to each channel of the image, where $c$ denotes the $c$-$th$ channel of the multi-channel image $I_{t-1}$. The Fourier transform can be expressed as:
\begin{equation}
\mathcal{F}(I_c)(u, v) = \frac{1}{\sqrt{HW}} \sum_{h=0}^{H-1} \sum_{w=0}^{W-1} I_c(h, w) e^{-j^{2\pi} \left( \frac{h}{H} u + \frac{w}{W} v \right)}
\end{equation}
where $u$ and $v$ denote the frequency-domain coordinates, $h$ and $w$ represent the spatial-domain coordinates. $j^2=-1$ is the imaginary unit. The magnitude and phase components in the frequency domain are computed as follows: 
\begin{equation}
\mathcal{A}(I_c)(u, v) = \sqrt{R^2(I_c)(u, v) + I^2(I_c)(u, v)}
\end{equation}
\begin{equation}
\mathcal{P}(I_c)(u, v) = \arctan\left[ \frac{I(I_c)(u, v)}{R(I_c)(u, v)} \right]
\end{equation}
where $R(I_c)$ and $I(I_c)$ denote the real and imaginary parts of the complex-valued representation, respectively. During the reverse denoising process, a soft fusion strategy is applied to the magnitude component, where a weighted averaging operation is performed in the Fourier domain to progressively align the magnitude with that of the original image. In contrast, the phase component is directly replaced with the phase of the original image obtained from the forward process. The corresponding operations on the magnitude and phase components are formulated as follows:
\begin{equation}
\tilde{I}_{t-1} = \mathcal{F}^{-1}\left( \Phi_\beta \mathcal{A}(\tilde{I}_{t-1}) + (1-\Phi_\beta) \mathcal{A}(I_{t-1}), \mathcal{P}(I_{t-1}) \right)
\end{equation}
where $\tilde{I}_{t-1}$ denotes the image at step $t$-$1$ in the forward diffusion process, $I_{t-1}$ denotes the image at step $t$-$1$ in the reverse diffusion process, and $\mathcal{F}^{-1}(\cdot)$ represent the inverse Fourier transform. Following \cite{Yang2020FDAFD}, $\Phi_\beta$ denotes a frequency-domain mask with nonzero values restricted to a small central region, and it is defined as $[-\beta H: \beta H, -\beta W: \beta W]$. The image center is aligned at (0, 0), and  $\beta \in (0, 1)$ controls the size of the preserved low-frequency region.

Finally, the fused magnitude and phase components are transformed back to the spatial domain via the inverse Fourier transform. To further restore fine-grained image details, a dynamic perturbation regulation mechanism is introduced. Specifically, stronger perturbations are applied in the early sampling stage to more effectively suppress watermark signals, while the perturbation strength is progressively reduced in later stages to better preserve image details and enhance visual fidelity. The dynamic perturbation regulation is formulated as follows:
\begin{equation}
L(t) = L_{\min} + (L_{\max} - L_{\min}) \cdot \frac{t}{T}
\end{equation}
where $t$ represents the current sampling step, $T$ denotes the maximum number of sampling steps,and $L_{\max}$ and $L_{\min}$ represent the maximum and minimum perturbation values, respectively.
\subsection{Phased Training Strategy}
In watermark attack task, high visual quality of the attacked images is essential. However, conventional watermark attacks often introduce noticeable texture degradation, leading to unsatisfactory perceptual fidelity. In addition, when diffusion models are optimized only with the noise estimation objective, they mainly learn to invert the noise distribution, which is insufficient to guarantee both faithful texture preservation and effective watermark suppression. Consequently, although the reconstructed outputs can be driven toward the distribution of watermark-free images, a favorable trade-off between visual fidelity and attack effectiveness remains difficult to achieve.
\begin{table*}[!t]
\centering
\caption{Comparison of Different Attack Methods Across Various Watermarking Methods on CelebA and ImageNet (Bold Indicates the Best Performance).}
\label{tab:psnr_attack}
\renewcommand{\arraystretch}{1.5}
\resizebox{\textwidth}{!}{
\begin{tabular}{>{\centering\arraybackslash}m{3cm} c|cccccccc}
\hline
\multicolumn{2}{c|}{} 
& \multicolumn{4}{c|}{CelebA (PSNR/BER)} 
& \multicolumn{4}{c}{ImageNet (PSNR/BER)} \\
\hline
\multicolumn{2}{c|}{\multirow{2}{*}{Attack methods}} 
& \multicolumn{8}{c}{Watermarking methods} \\
\cline{3-10}
\multicolumn{2}{c|}{} 
& LSB \cite{faheem2023edge}& DCT \cite{cao2024universal}& QPHFMs \cite{He2024ExploringAI}& \multicolumn{1}{c|}{HiDDeN\_MP \cite{DBLP:conf/aisec-ws/HuangYHSCC25}} 
& LSB \cite{faheem2023edge}& DCT \cite{cao2024universal}& QPHFMs \cite{He2024ExploringAI}& HiDDeN\_MP \cite{DBLP:conf/aisec-ws/HuangYHSCC25}\\
\hline

\multicolumn{1}{>{\centering\arraybackslash}m{3cm}|}{\multirow{5}{*}{Traditional}} 

& Gaussian noise (0.002) 
& 27.16/0.3125 & 27.16/0.1916 & 27.02/0.0859 & \multicolumn{1}{c|}{26.98/0.0563} 
& 27.28/0.3174 & 27.29/0.1979 & 28.24/0.0469 & 26.35/0.0674 \\

\multicolumn{1}{>{\centering\arraybackslash}m{3cm}|}{} 

& Speckle noise (0.005) 
& 28.78/0.2949 & 28.77/0.1298 & 30.31/0.0352 & \multicolumn{1}{c|}{29.46/0.0708} 
& 28.72/0.3096 & 28.73/0.1378 & 34.40/0.0195 & 29.63/0.0523 \\

\multicolumn{1}{>{\centering\arraybackslash}m{3cm}|}{} 

& Salt \& pepper noise (0.005) 
& 27.97/0.3215 & 27.93/0.1621 & 28.75/0.0664 & \multicolumn{1}{c|}{28.14/0.0963} 
& 27.88/0.3320 & 27.96/0.1525 & 27.04/0.0703 & 27.78/0.0738 \\

\multicolumn{1}{>{\centering\arraybackslash}m{3cm}|}{} 

& Average filter (3$\times$3) 
& 30.40/0.2513 & 30.36/0.2734 & 32.09/0.1680 & \multicolumn{1}{c|}{33.86/0.1876} 
& 25.96/0.2636 & 25.96/0.2729 & 28.09/0.2383 & 29.22/0.2153 \\

\multicolumn{1}{>{\centering\arraybackslash}m{3cm}|}{} 

& JPEG compression 30 
& 33.74/0.2845 & 33.68/0.2668 & 34.43/0.1625 & \multicolumn{1}{c|}{34.26/0.1963} 
& 29.58/0.3216 & 29.52/0.2844 & 29.52/0.2383 & 29.58/0.3014 \\

\hline

\multicolumn{1}{>{\centering\arraybackslash}m{3cm}|}{\multirow{3}{*}{Deep learning}} 

& RD-IWAN \cite{Wang2022RDIWANRD}
& 46.24/0.2765 & 37.32/0.2945 & 41.65/0.0758 & \multicolumn{1}{c|}{42.33/0.2765} 
& 43.99/0.3142 & 36.24/0.3059 & 39.57/0.0963 & 40.67/0.2895 \\

\multicolumn{1}{>{\centering\arraybackslash}m{3cm}|}{} 

& DiffWA  \cite{Li2023DiffWADM} 
& 30.98/0.2894 & 30.47/0.1692 & 30.83/0.1687 & \multicolumn{1}{c|}{31.84/0.4341} 
& 29.46/0.3158 & 28.76/0.1793 & 28.86/0.1452 & 29.89/0.4427 \\

\multicolumn{1}{>{\centering\arraybackslash}m{3cm}|}{} 

& HIWANet \cite{Wang2024HIWANetAH}
& 32.81/0.3156 & 30.76/0.2534 & 31.55/0.1794 & \multicolumn{1}{c|}{31.75/0.2649} 
& 31.32/0.3281 & 30.48/0.2947 & 30.04/0.1922 & 29.77/0.2934 \\

\hline

\multicolumn{1}{>{\centering\arraybackslash}m{3cm}|}{Proposed} 

& \textbf{FMDiffWA} 
& \textbf{45.26/0.3203} & \textbf{43.41/0.3386} & \textbf{42.18/0.2031} & \multicolumn{1}{c|}{\textbf{43.63/0.4495}} 
& \textbf{43.82/0.3327} & \textbf{40.16/0.3415} & \textbf{40.45/0.2264} & \textbf{41.53/0.4562} \\

\hline
\end{tabular}
}
\end{table*}
As described in Section 4.2, we adopt a stage-wise training strategy to mitigate the limitations of relying solely on the noise estimation objective. Notably, the loss functions in both stages are computed on fixed-size image patches. In the initial stage, training is dominated by the noise estimation constraint in Eq. \eqref{eq:eq7}, which enables the U-Net to accurately predict the noise  at each time step. Based on Eq.  \eqref{eq:eq6}, the corresponding mean and variance of $I_{t-1}$ can be derived, which helps maintain the stability of the sampling process. Subsequently, a sampling-based refinement constraint is introduced in the middle and later stages. During this refinement stage, the model first produces a coarse restored image from early-stage sampling and then learns to align it with the watermark-free image. Through this progressive optimization process, the model gradually generates refined results with higher visual quality while more effectively disrupting the watermark signals. In this stage, the refinement constraint is implemented using an $\mathcal{L}_1$ loss and a multi-scale structural similarity loss. According to existing studies \cite{DBLP:journals/tci/ZhaoGFK17}, compared with $\mathcal{L}_2$ loss, $\mathcal{L}_1$ loss yields better overall performance in terms of convergence speed and reconstruction accuracy. Meanwhile, the MS-SSIM loss, denoted as $\mathcal{L}_{\text{ms-ssim}}$, emphasizes structural consistency between the generated images and the original image patches from a multi-scale perspective, and is defined as follows:
\begin{equation}
\mathcal{L}_{\text{ms-ssim}} = 1 - \prod_{j=1}^{M} \left( \frac{2 u_{\tilde{I}_a^{(i)}} u_{I^{(i)}} + C_1}{u_{\tilde{I}_a^{(i)}}^2 + u_{I^{(i)}}^2 + C_1} \cdot \frac{2 \sigma_{\tilde{I}_a^{(i)} I^{(i)}} + C_2}{\sigma_{\tilde{I}_a^{(i)}}^2 + \sigma_{I^{(i)}}^2 + C_2} \right)
\end{equation}
where $u_{\tilde{I}_a^{(i)}}$,  $u_{I^{(i)}}$, $\sigma_{\tilde{I}_a^{(i)}}$ and $\sigma_{I^{(i)}}$ denote the means and standard deviations of $\tilde{I}_a^{(i)}$ and $I^{(i)}$, respectively. $\sigma_{\tilde{I}_a^{(i)} I^{(i)}}$ represents the covariance between them. $C_1$ and $C_1$ are constant values, which are typically set to $1 \times 10^{-4}$ and $9 \times 10^{-4}$. $M$ denotes the number of scales and is set to 5 following \cite{Wang2003MultiscaleSS}. The overall refinement constraint is defined as follows:
\begin{equation}
\mathcal{L}_{\text{ref}} = \mathcal{L}_1 + \mathcal{L}_{\text{ms-ssim}}
\label{eq:loss_ref}
\end{equation}
\section{Experiments}
In this section, the implementation details of the experiments are first presented. The proposed method is then compared with existing approaches on the CelebA and ImageNet datasets in terms of both the visual quality and the watermark attack ability. Ablation studies are subsequently conducted to validate the effectiveness of each component of the proposed model.

\subsection{Experimental Setting}
Datasets. Experiments are conducted on the face-oriented CelebA dataset and the general-purpose natural image dataset ImageNet. All images are resized to $256\times256$. A total of $24000$ images are randomly selected, and a $16\times16$ watermark is embedded into each image using four representative watermarking algorithms. Among these images, $18000$ images are used for training, and the remaining $6000$ images are reserved for performance evaluation. The four watermarking methods include the spatial-domain-based LSB \cite{faheem2023edge} method, the frequency-domain-based DCT \cite{cao2024universal} method, the orthogonal moment-based PHFMs \cite{He2024ExploringAI} method, and the deep learning-based \text{HiDDeN\_MP} \cite{DBLP:conf/aisec-ws/HuangYHSCC25} method.

Training Details. All experiments are implemented in the PyTorch 2.1.0 and trained on an NVIDIA RTX 4090 GPU. During training of the FMW module, the Adam optimizer is adopted with $\beta_1=0.9$ and $\beta_2=0.999$. The batch size is set to 2, and the initial learning rate is $2e$-$5$. Each input image is randomly cropped into 16 patches of $64\times64$, which are then fed into the diffusion model. The model is first trained for $1\times10^6$ iterations with the noise estimation objective, followed by an additional $3\times10^5$ iterations with the refinement constraint. The total number of diffusion time steps $T$ is set to 1000, and the number of implicit sampling steps is set to 10. In addition, an exponential moving average (EMA) strategy is applied to reduce color shifts and artifacts in the output images, thereby improving the visual imperceptibility of the attacked results.

\subsection{Watermark Removal Ability Evaluation}
In this section, FMDiffWA is compared with existing watermark attack methods. The compared methods are broadly divided into traditional attacks and deep learning–based attacks. Specifically, the traditional attacks include Gaussian noise with a variance of 0.002, speckle noise with a density of 0.005, salt-and-pepper noise with a density of 0.005, average filtering with a $3\times3$ window, and JPEG compression with a quality factor of 30. The deep learning–based baselines include RD-IWAN \cite{Wang2022RDIWANRD}, DiffWA \cite{Li2023DiffWADM}, and HIWANet \cite{Wang2024HIWANetAH}. To ensure a fair comparison, traditional attacks, which do not require additional training, are directly applied to the watermarked images. For deep learning–based watermark attacks that require training, all comparison models are retrained on the CelebA and ImageNet datasets according to the implementation details reported in their original papers before evaluation.
\begin{figure}[htbp]
  \centering
  \includegraphics[page=3,width=1\linewidth]{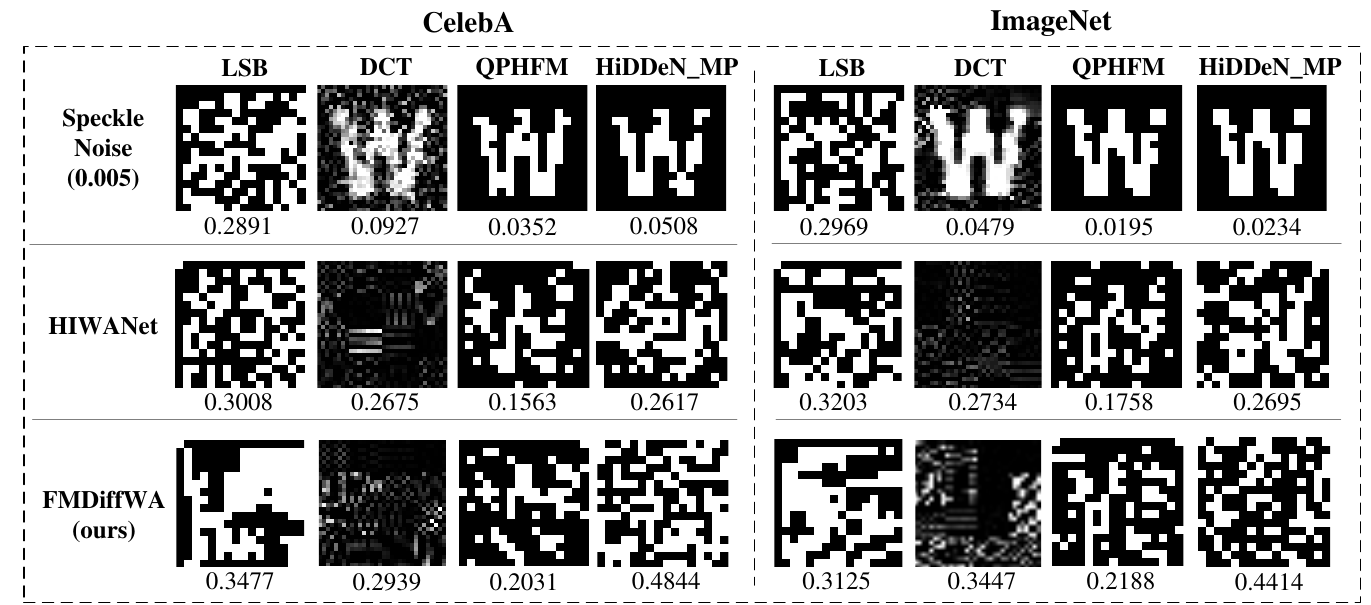}
  \caption{Visualization of watermark images extracted from CelebA and ImageNet images after attacks by speckle noise, HIWANet, and the proposed FMDiffWA, together with the corresponding BERs.}
  \label{Figure:Figure3}
\end{figure}

\begin{figure*}[!t]
    \centering
    \includegraphics[page=4,width=0.9\textwidth]{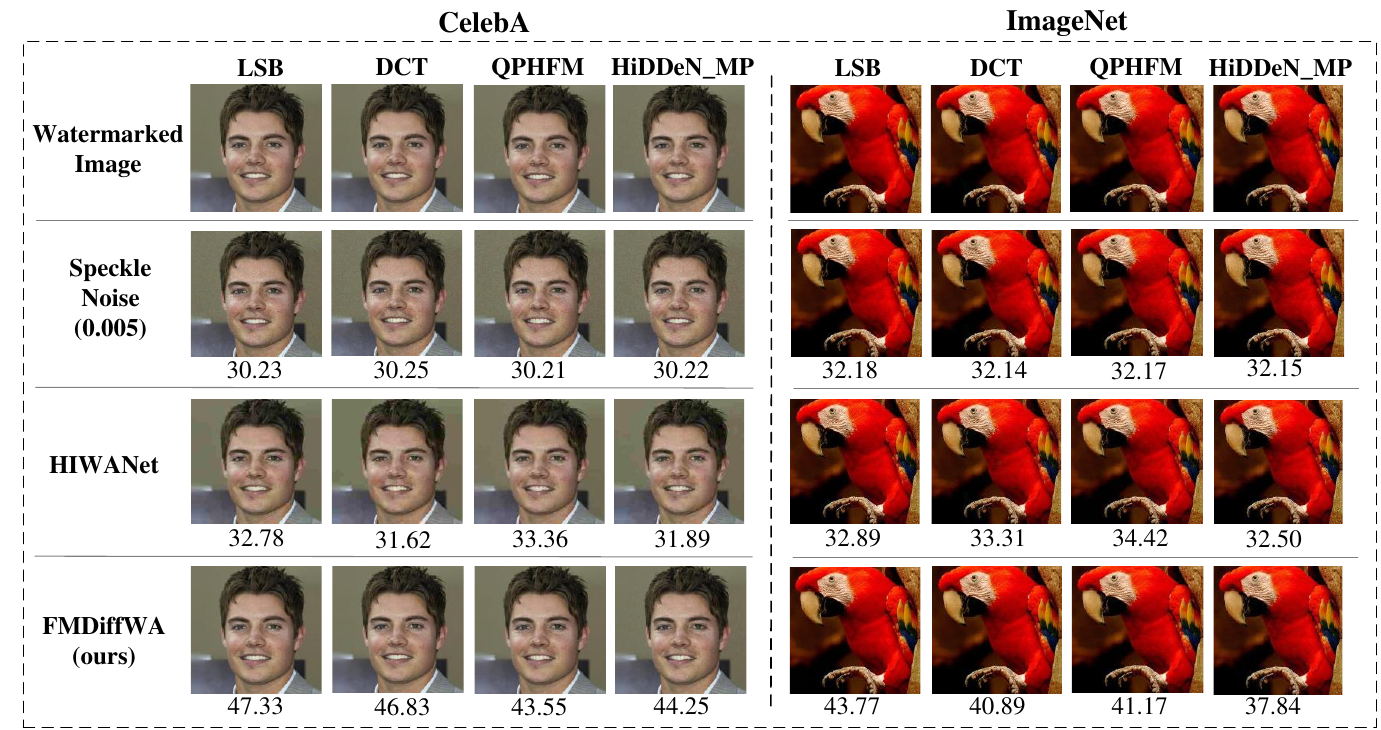}
    \caption{Visualization of watermark attacks on CelebA and ImageNet under Speckle noise, HIWANet, and the proposed FMDiffWA.}
    \label{fig:Figure4}
\end{figure*}
The experimental results in Table \ref{tab:psnr_attack} demonstrate that FMDiffWA consistently achieves superior watermark attack performance across different watermarking algorithms. In contrast, the traditional watermark attacks generally produce much lower BERs\cite{Ali2013BitErrorRateS}, especially when attacking robust watermarks using QPHFMs and deep learning-based watermarking algorithms. This limitation mainly arises because traditional attacks rely on blind perturbations, and their effectiveness is highly sensitive to parameter settings. As a result, extensive manual tuning is often required for different watermarking algorithms, which makes it difficult to obtain consistently strong attack performance across diverse watermarking schemes.

Furthermore, existing deep learning-based attacks typically operate directly in the spatial domain, and learn end-to-end pixel-level representations. Such methods often struggle to effectively disrupt watermark details in the frequency domain. By contrast, FMDiffWA introduces FWM module that incorporates magnitude and phase information from the forward diffusion process, thereby enabling more effective attack of watermark information. Through the combination of FWM module and diffusion modeling, FMDiffWA can effectively disrupt watermarks while restoring high-quality images with strong visual fidelity.

\begin{figure}[htbp]
  \centering
  \includegraphics[width=1\linewidth]{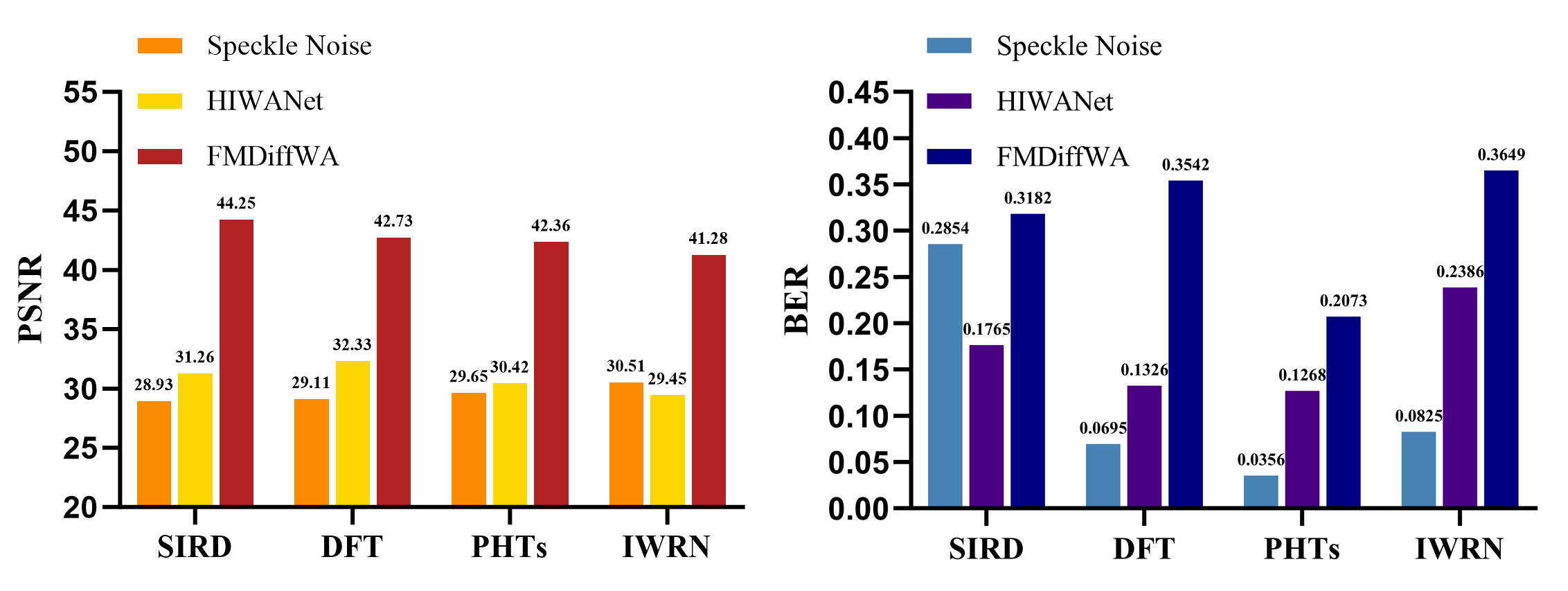}
  \caption{Comparison of generalization performance to unseen watermarking methods under Speckle noise, HIWANet, and the proposed FMDiffWA.}
  \label{Figure:Figure5}
\end{figure}

The results show that, after multiplicative noise attack, the watermarks by different methods can still be successfully extracted. In particular, the PHFMs-based watermarking method is almost unaffected by Gaussian noise, indicating strong robustness. When deep learning-based attacks are applied, HIWANet exhibits stronger attack capability and can effectively degrade watermarks embedded by multiple watermarking algorithms; however, the watermarks embedded by the QPHFMs method can still be extracted with relatively high accuracy. In contrast, FMDiffWA consistently achieves the best performance across all evaluated watermarking methods, making the watermark information completely unrecoverable after the attack. These results clearly demonstrate that FMDiffWA possesses outstanding watermark attack ability and strong generalization capability against different watermarking algorithms.

\subsection{Visual Quality Comparisons}
In this section, the visual quality of images attacked by FMDiffWA is evaluated and compared with that of other watermark attacks across multiple watermarking algorithms. The quantitative results are reported in Table \ref{tab:psnr_attack}. As shown in Table \ref{tab:psnr_attack}, traditional attacks cause significant degradation in image quality, with PSNR \cite{Yuanji2003ImageQE} values remaining around 30 dB after attacks. For deep learning-based attacks, the PSNR values of the attacked images generally exceed 30 dB. In contrast, FMDiffWA achieves higher visual fidelity, with PSNR values consistently above 40 dB after attacks. These results demonstrate that FMDiffWA significantly outperforms existing attacks in terms of preserving visual quality while effectively removing watermarks.

As shown in Table \ref{tab:psnr_attack}, traditional attacks typically cause significant degradation in visual quality. These attacks mainly rely on signal processing operations such as JPEG compression and spatial filtering, which introduce global perturbations accross the entire image, resulting in noticeable distortions and reduced PSNR values. By jointly exploiting the forward and reverse processes of diffusion models, FMDiffWA can disrupt the embedded watermarks while accurately recovering the watermark-free image. In addition, a stage-wise training strategy is adopted, in which a refinement constraint is introduced after the noise estimation objective. As a result, FMDiffWA consistently achieves the highest PSNR values across all experiments, which demonstrates its superior capability for preserving image quality during watermark attack.
\begin{figure}[!t]
    \centering
    \includegraphics[width=1\linewidth]{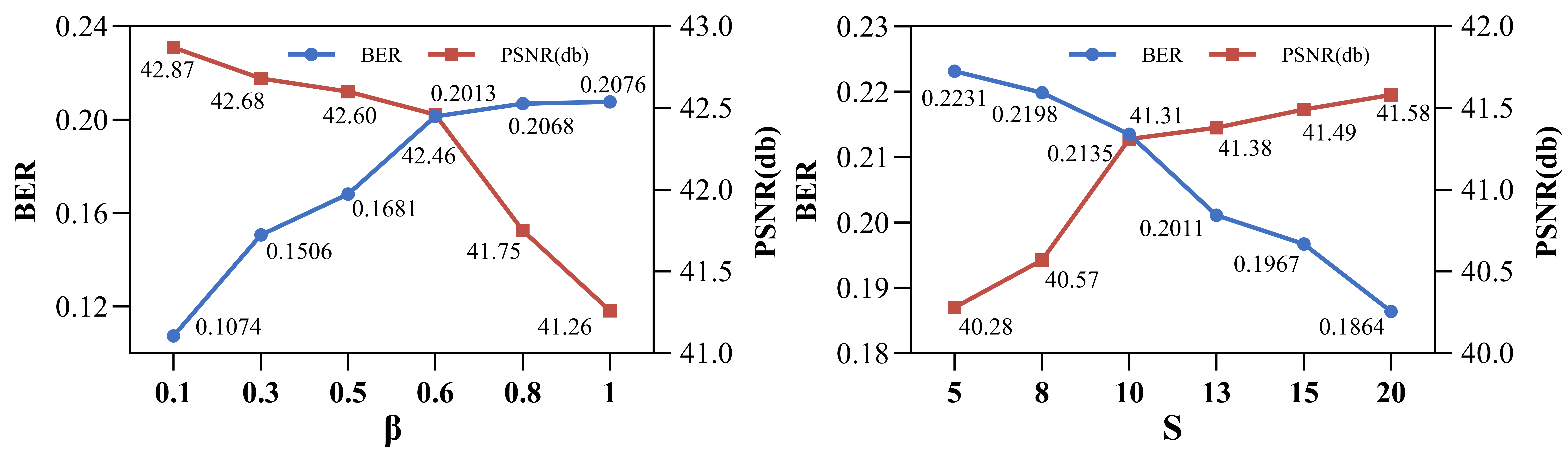}
  \caption{Comparison of the watermark removal ability and image quality of FMDiffWA under varying hyperparameters $\beta$ and $S$.}
  \label{Figure:Figure6}
\end{figure}

In addition, Fig. \ref{fig:Figure4} provides a visual comparison among images attacked by Speckle Noise, HIWANet, and FMDiffWA. It can be clearly observed that the images processed by FMDiffWA exhibit the smallest perceptual difference from the corresponding watermarked images. This result further indicates that FMDiffWA can effectively destroy watermark information while preserving a high degree of visual consistency. Owing to its strong watermark attack ability and superior visual fidelity, FMDiffWA shows considerable potential for practical watermark attack applications.

\subsection{Generalization Evaluation}
Generalization is a critical property of watermark attacks, as it reflects the ability of an attack to maintain stable performance across different watermarking methods. A watermark attack with strong generalization capability can effectively destroy watermarks embedded by different watermarking algorithms, which is essential for the practical applicability of watermark attacks in real-world scenarios. The generalization performance of FMDiffWA is evaluated under four different watermarking algorithms, namely SIRD \cite{Abraham2016AnIS}, DFT \cite{urvoy2014perceptual}, PHTs \cite{Tang2024ARR}, and IWRN \cite{Kou2025IWRNAR}. The experiment uses 400 images with a resolution of $256\times256$ and a single $16\times16$ binary watermark. For every 100 images, watermark embedding is performed using the same watermarking algorithm, resulting in a total of 400 watermarked images.
\begin{table}[h]
\centering
\caption{Ablation Study of Key Components for Watermark Attacks on Images with QPHFMs Watermarks.}
\label{tab:ablation_modules}
\begin{tabular}{cc cc}
\toprule
\multicolumn{2}{c}{Module} & \multirow{2}{*}{PSNR (dB)} & \multirow{2}{*}{BER } \\
\cmidrule(r){1-2}
FWM & PTS &  &  \\
\midrule
$\checkmark$ &  & 39.96 & 0.1896 \\
 & $\checkmark$ & 48.65 & 0.0664 \\
$\checkmark$ & $\checkmark$ & 42.08 & 0.2084 \\
\bottomrule
\end{tabular}
\end{table}
As illustrated in Fig. \ref{Figure:Figure5}, the comparison among speckle noise, HIWANet, and FMDiffWA shows that FMDiffWA consistently achieves effective watermark destruction across various unseen watermarking algorithms, thereby further validating its strong generalization capability and stable attack performance. By comparing the HIWANet method with our proposed FMDiffWA method, it can be clearly observed that FMDiffWA is still able to effectively destroy watermark information under various unseen watermarking algorithms, which further demonstrates its excellent generalization ability and stable attack performance. These results indicate that FMDiffWA captures the intrinsic characteristics of watermark embedding in images rather than overfitting to a specific watermarking algorithm. Consequently, when applied to different watermarking methods, the attack performance of FMDiffWA remains largely consistent, maintaining strong watermark destruction capability while preserving high image quality after attack.

\subsection{Ablation Study}
The Effectiveness of different componets. We adopt a block-based conditional DDPM as the baseline and evaluate different configurations using PSNR and BER. As shown in Table \ref{tab:ablation_modules}, we conclude that: (1) Without staged training, watermark removal improves but image quality degrades. (2) Without the FWM, the model preserves image quality but fails to remove watermarks. (3) Combining both modules achieves the best performance, effectively removing watermarks while maintaining high image quality.

Effect of Mask Scale in the FWM. The mask scale (hyperparameter $\beta$) controls the relative contribution of amplitude and phase components in the frequency-domain watermark modulation process. As shown in Fig. \ref{Figure:Figure6}, we vary $\beta$ from 0.1 to 1. The results show that $\beta=0.6$ achieves the best trade-off, effectively removing watermarks while preserving image quality.

Effect of Sampling Steps. During sampling, the model progressively generates clean images through $S$ denoising steps. Increasing $S$ generally improves visual quality \cite{Whang2021DeblurringVS} but also increases inference cost. We evaluate $S$ from 1 to 20. As shown in Fig. \ref{Figure:Figure6}, PSNR slightly decreases when $S > 10$, likely due to over-smoothing caused by excessive iterations. Considering both performance and efficiency, we set $S=10$ in our experiments.

\section{Conclusion}
In this paper, a watermark attack based on frequency-domain modulated diffusion (FMDiffWA) is proposed. FMDiffWA provides a new perspective on watermark attack by exploiting the generative capability of conditional diffusion models to improve the visual quality of attacked images. Unlike conventional conditional diffusion models, we introduce a frequency-domain modulation (FWM) module that guides the sampling process toward watermark-free images, thereby enabling more effective watermark suppression. In addition, a stage-wise training strategy is designed to address the suboptimal generation quality of diffusion models by introducing a refinement constraint after the initial noise estimation objective. Experimental results demonstrate that FMDiffWA significantly outperforms traditional watermark attacks and existing deep learning–based attacks. 

Future work will focus on improving the generalization of diffusion-based watermark attacks under more complex and diverse watermarking algorithms. In particular, it would be an important direction to explore stronger frequency-aware priors, more efficient sampling strategies, and better trade-offs between attack effectiveness and visual fidelity.


\bibliographystyle{ACM-Reference-Format}
\bibliography{main}


\appendix

\end{document}